\documentclass[11pt,a4paper]{article}
\usepackage[hyperref]{acl2021}
\usepackage{times}
\usepackage{latexsym}

\usepackage{microtype}

\usepackage{graphicx}

\aclfinalcopy 
\def\aclpaperid{2899} 

\title{Quantifying and Avoiding Unfair Qualification Labour in Crowdsourcing}

\author{Jonathan K. Kummerfeld \\
Computer Science \& Engineering \\
University of Michigan, Ann Arbor \\
\texttt{jkummerf@umich.edu}
}

\date{}

\begin{document}

\maketitle

\begin{abstract}

Extensive work has argued in favour of paying crowd workers a wage that is at least equivalent to the U.S.~federal minimum wage.
Meanwhile, research on collecting high quality annotations suggests using a qualification that requires workers to have previously completed a certain number of tasks.
If most requesters who pay fairly require workers to have completed a large number of tasks already then workers need to complete a substantial amount of poorly paid work before they can earn a fair wage.
Through analysis of worker discussions and guidance for researchers, we estimate that workers spend approximately 2.25 months of full time effort on poorly paid tasks in order to get the qualifications needed for better paid tasks.
We discuss alternatives to this qualification and conduct a study of the correlation between qualifications and work quality on two NLP tasks.
We find that it is possible to reduce the burden on workers while still collecting high quality data.

\end{abstract}

\section{Introduction}

Workers using Amazon Mechanical Turk earn a median wage of \$2.54 an hour \cite{earnings}, far below the U.S.-federal minimum wage of \$7.25.
Many researchers pay workers a higher wage, estimating the time spent on a task and giving bonuses when the time required is higher than expected.
At the same time, researchers try to maintain the quality of work completed using a variety of methods \cite{person-centric}.
One common approach, used by 19\% of tasks (HITs) on the platform \cite{earnings}, is to restrict tasks to workers who have had a certain number of HITs approved.
Tasks with this restriction have a median wage of \$4.14 an hour, far above the overall average.
If most high paying requesters use this restriction it means workers need to do a substantial amount of low paid ``Qualification Labour": work to achieve the qualifications necessary for fairly paid tasks.
These tasks may also be particularly unpleasant work that more experienced workers are unwilling to do, e.g., they might involve unsavoury content.

This paper is the first to identify the qualification labour issue and explore it.
We study norms around the setting of the qualification and the effort workers put in to achieve common milestones.
5,000 accepted tasks, a common requirement, takes over 2 months of effort.
We consider several ways to address the issue, and study the work quality of groups with different qualifications.\footnote{Code for our experiments is attached to this paper}
Using two tasks, coreference resolution and sentiment analysis, we find that high quality annotations can be collected with a lower threshold, though there are task dependent patterns.

\section{Background and Related Work}

Crowd work involves large groups of workers doing small paid tasks, known as Human Intelligence Tasks (HITs).
Services such as Amazon Mechanical Turk provide a marketplace to connect workers with requesters.
Requesters create tasks, workers choose which tasks to do, then either complete them or return them.
Requesters approve or reject the completed work.
Tasks can be restricted to workers with certain qualifications, e.g.~location.
Amazon tracks some statistics that can be used as qualifications.
This work focuses on (1) the total number of approved HITs a worker has, and (2) the percentage of their HITs that were accepted.

Since the earliest uses of crowd work in NLP, there has been work discussing issues such as poor wages and the lack of worker rights \cite{fort-etal-2011-last}.
These have also been discussed in the Human-Computer Interaction research community \cite{10.1145/1979742.1979606,earnings}.
There has been work on proposing guidelines for requesters \cite{sabou-etal-2014-corpus}, incorporating workers into the IRB process \cite{neurips20}, and developing tools to help workers address the power imbalance in the online workplace \cite{10.1145/2470654.2470742,10.1145/2858036.2858592}.
Concurrent with this work, another study showed that crowdsourcing is being used more each year in NLP research, and there is limited awareness of the ethical issues in this type of work \cite{shmueli-etal-2021-beyond}.

Prior work has considered hidden labour in the day-to-day work of the crowd \cite{earnings}.
By observing a large set of workers, they measured time involved in searching for tasks, returned tasks, and breaks.
Some of these issues have received additional attention, such as the wasted effort on tasks that are returned rather than completed \cite{10.1145/3289600.3291035}.
While informative, those studies do not account for the hidden labour identified in this paper, which spans a long period and relates to worker qualifications.

Part of this work uses online discussion between workers to understand their work.
Prior work has used a similar approach to understand the overall experience of crowd workers \cite{10.1145/2531602.2531663}.

\section{Norms for the Approved HITs Value}

The value used as the Approved HITs threshold is rarely reported in prior work.
Three recent papers specify a 1,000 HIT threshold \cite{hcomp2019,creativity,hcomp-fairwork}.
Outside of Computer Science, advice in articles \cite{social-science-pub} and tutorials \cite{melb-tutorial} is to set the value to 100 because that is when another qualification (approval percentage) becomes active.
This difference may be because other fields primarily use crowdsourcing for surveys rather than data annotation or human computation systems.
It is unclear how representative these samples are.
However, there are other sources that can provide information about conventions.

One source is Amazon itself.
The Mechanical Turk web-interface provides six options: 50, 100, 500, 1,000, 5,000, 10,000.
The MTurk blog has mentioned this qualification in four posts over the past eight years \cite{mturk-blog0,mturk-blog1,mturk-blog2,mturk-blog3}.
In three cases, the value was 5,000 and in the fourth it was 10,000.

Another source is forums and blogs.
One pinned thread on the MTurk Crowd forum advises that ``For your first 1000 HITs you may want to concentrate on approval milestones rather than \$\$\$ ... most of the better-paying requesters require 1000/5000/10000+ approved HITs" \cite{crowd-form0}.
This advice is repeated elsewhere on the forum and on Reddit \cite{crowd-form1,reddit4,reddit5}.
This is consistent with observations that 80\% of tasks available to new users pay less than 10 cents \cite{fair-share}.
In one discussion between a worker and a requester, the worker recommended a threshold of 5,000 \cite{crowd-form2}.
In the blog ``Tips For Requesters On Mechanical Turk", one post recommends at least 5,000 if not 10,000 \cite{turk-requesters0} while another recommends at least 1,000 \cite{turk-requesters1}.
A web article by a Computer Vision researcher recommended 1,000 \cite{kumar}.
The CloudResearch blog mentions the threshold once, noting that a value of 10,000 maintains quality without significantly increasing the time to finish a set of HITs \cite{cloud-research0}.

Qualifications are also discussed by courses and tutorials.
In the Crowdsourcing \& Human Computation course at the University of Pennsylvania, a guest lecture on ``The Best Practices of the Best Requesters" mentioned the approved HITs qualification and used 10,000 as an example \cite{upenn}.
One guide recommends a cutoff of 5,000 \cite{ucsd}.

Overall, we conclude that while practises vary, 5,000 or higher are commonly used as a qualification for tasks.

\subsection{Impact on Workers}

It is difficult to estimate how much time workers have to spend to achieve this qualification.
Academic studies of time spent on HITs may be skewed by experienced workers, who have strategies for finding and completing tasks rapidly.
Posts on Reddit mention taking anywhere from a month to a year to reach 5,000 approved HITs.
The median of values reported across several Reddit threads was 2.25 months \cite{reddit0,reddit1,reddit2,reddit3}.
Assuming 20 hours of work a week that is almost 200 hours of effort (140 seconds per task).

\subsection{Potential Solutions}

If this type of qualification undercuts our commitment to paying a fair wage, what are alternative ways to maintain quality?
Options include:

1. Introduce screening questions that workers must complete correctly to proceed to the rest of the task, e.g., requiring 70\%+ on three questions \cite{hcomp-screen}.
This approach is problematic because it the workers who fail the screening are doing unpaid labour.

2. Address quality after collection by either dropping the lowest performing workers \cite[e.g., the bottom 25\% in][]{hcomp-mental-models}, aggregating a larger number of responses per example, or including attention check questions and discarding workers who get them wrong.
All of these incur a substantial cost to researchers.

3. Controlled crowdsourcing \cite{roit-etal-2020-controlled} uses an initial task that a broad set of workers can complete and then limits participation to the workers who did well on that task.\footnote{
One potential drawback of this approach is that the filtering step may produce a biased sample of workers.
That may be problematic for more subjective tasks, though with a large enough sample, responses could be weighted to make the results more representative.
}
The cost of this solution depends on the percentage of workers who do well on the initial task.

4. Lower the threshold, reducing the required volume of earlier work.
This reduces, but does not eliminate the qualification labour issue.

These methods can also be combined.
Controlled crowdsourcing (method 3) with a very low Accepted HITs threshold (method 4) for the initial task would address the ethical concern we raise here while limiting the additional cost to the recruitment phase. 
Attention checks and aggregation (method 2) would then address natural variation in skill and attention during large-scale annotation.

\section{Studying the Approved HITs Value}

All of the options above have tradeoffs that will be task dependent and in practise some combination is most likely to be the best approach.
The first three have been studied in prior work, but the impact of lowering the threshold has not.
In this section, we consider the quality of work completed by workers grouped by how many tasks they have previously completed and what percentage were accepted.\footnote{
This was completed as part of a larger study approved by the Michigan IRB under study ID HUM00155689.
}

\subsection{Tasks}

\begin{figure}
    \centering
    \includegraphics[width=\linewidth]{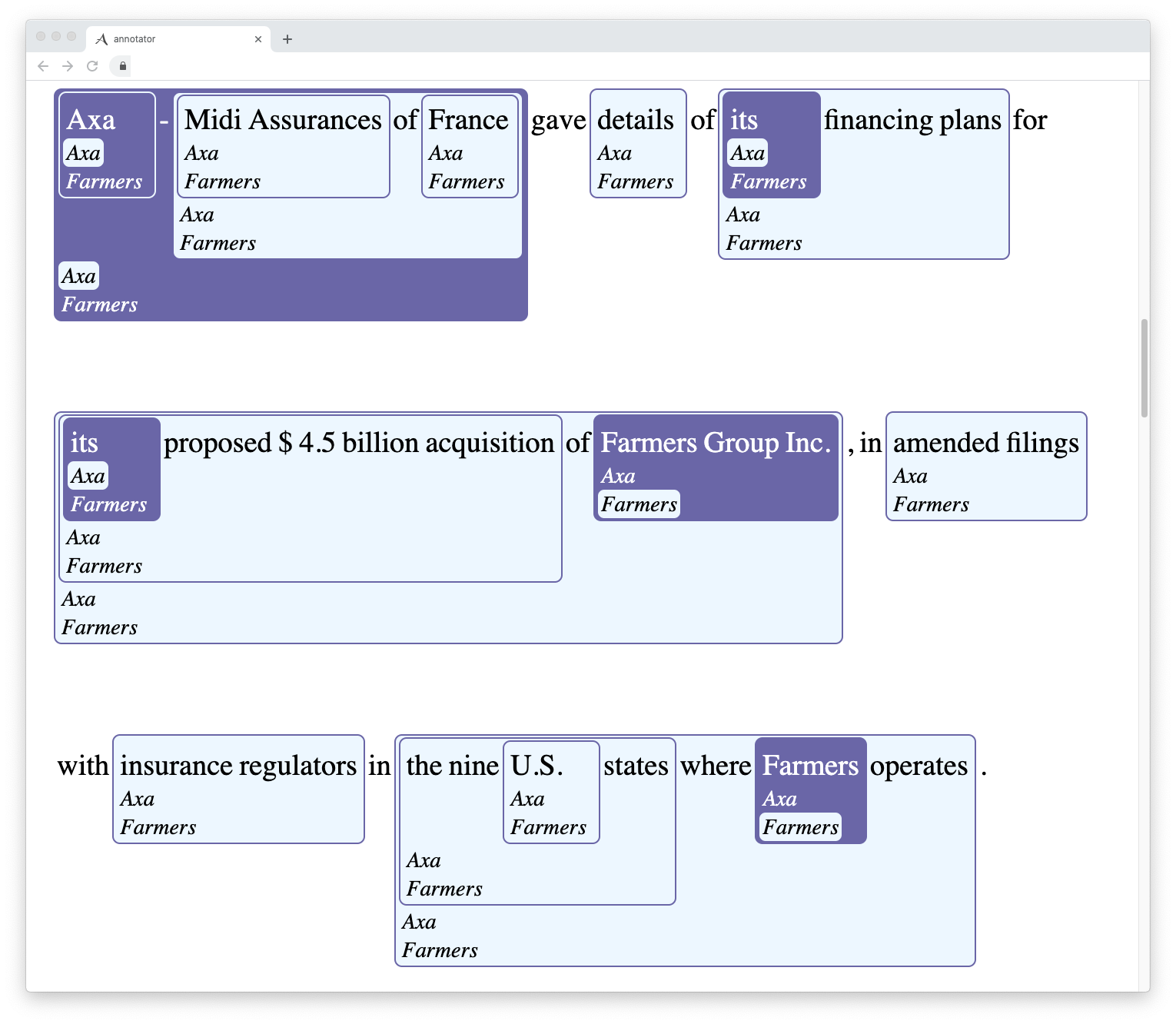}
    \caption{\label{fig:ui}
    The user interface for coreference resolution (zoomed in).
    Spans are noun phrases automatically assigned by the Allen NLP syntactic parser \cite{gardner-etal-2018-allennlp}.
    The two entities being identified are the two most frequently mentioned entities in the text.
    Workers select a label by clicking on it.
    }
\end{figure}

\paragraph{Coreference Resolution}
This is an unusual task for crowdsourcing, with a novel user interface, shown in Figure~\ref{fig:ui}.
Workers were shown a 244 word document from the Ontonotes dataset \cite{pradhan-xue-2009-ontonotes}.
We identified noun phrases using the Allen NLP parser \cite{gardner-etal-2018-allennlp} and asked workers to identify when one of two specific entities was mentioned.
This is not the complete coreference resolution task, but a useful subset.
We refined the task over several rounds of trial annotation to ensure the instructions were clear and the interface was efficient.
Workers were asked to check their answers if they tried to submit in less than 75 seconds.
If they labeled 8 items in the first 19 words, they were reminded to only label the two entities specified.
We estimated that the task would take 3 minutes and paid workers 60 cents (\$12 / hour).
Reviews on TurkerView (\url{https://turkerview.com/}) indicated that workers effective hourly rates were \$7.88, \$11.25, \$12.93, and \$14.59.

To evaluate performance, we calculated the F-score using the original Ontonotes annotations as the gold standard.
A score of 80\% or above was considered a good effort, to allow for minor errors and points of confusion.

\begin{figure*}
    \centering
    \includegraphics[width=0.9\linewidth]{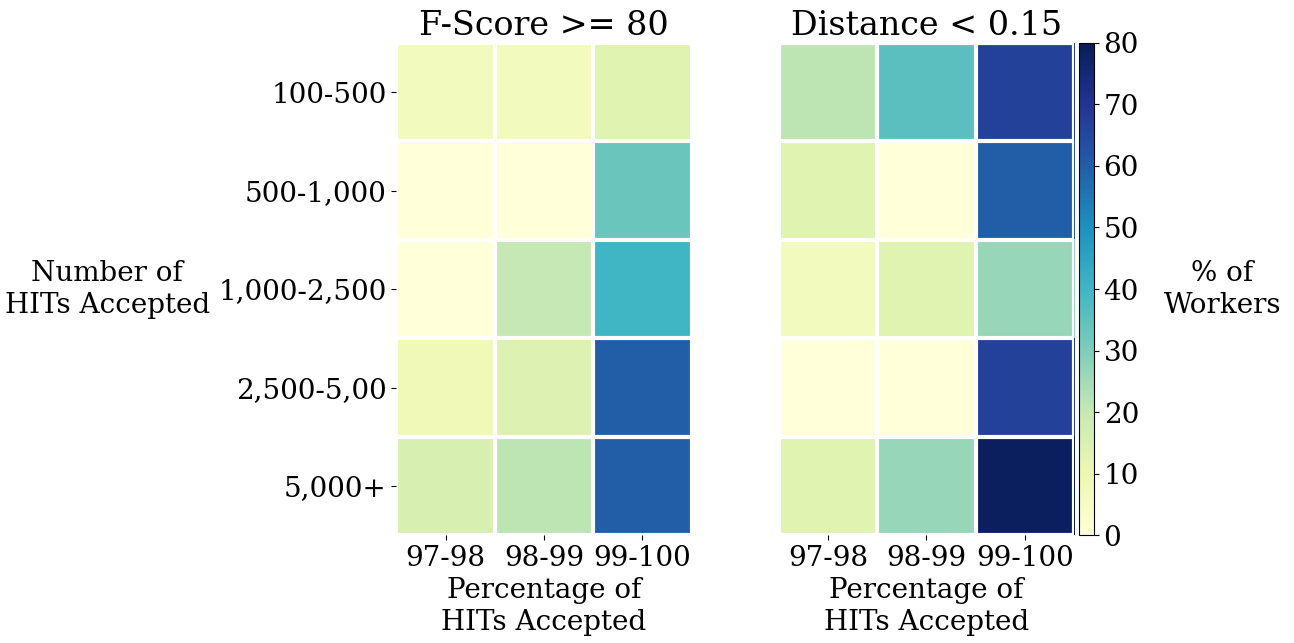}
    \caption{\label{fig:acc}
    Results for all fifteen combinations of qualifications.
    Left (coreference): The percentage of workers scoring above 80 in each group.
    Right (sentiment): The percentage of workers whose average error was below 0.15 in each group.
    Each value is based on fifteen workers, except
    for sentiment there were fourteen for (98-99\%, 500-1,000),
    and for coreference there were fourteen for (97-98\%, 500-1,000), (97-98\%, 1,000-2,500), (98-99\%, 2,500-5,000), (98-99\%, 5,000+),
    thirteen for (97-98\%, 5,000+),
    and twelve for (97-98\%, 2,500-5,000).
    }
\end{figure*}

\paragraph{Sentiment Analysis}
This task is very intuitive and has been crowdsourced extensively in the past.
We closely followed the set up used to annotate the Stanford Sentiment Treebank \cite{socher-etal-2013-recursive}, with the same task instructions.
Workers were shown ten examples whose true scores were evenly spread across 0 to 1.
We estimated that the task would take 4 minutes and paid workers 80 cents (\$12 / hour).
Three reviews of the task on TurkerView indicated that workers hourly earnings were \$22.15, \$48.00, and \$50.53, suggesting that workers were faster than anticipated.

To evaluate, the labels assigned are mapped to a scale from 0 to 1 and compared with the values in the original dataset.
An average value of below 0.15 was considered a reasonable score.
This cutoff was chosen based on the scores achieved by two NLP students in our lab (0.11 and 0.09).

\subsection{Recruitment}
We considered 15 combinations of ranges for ``Approved HITs" and ``Percentage Approved", as shown by the axis labels in Figure~\ref{fig:acc}.
The ranges are based on the preset values provided by MTurk, with the addition of a boundary at 2,500 to provide slightly more detail in the shift between 1,000 and 5,000.
Workers also had to be U.S.-based.
We used Javascript-based checks to ensure each worker completed the task only once.
224 workers completed the sentiment task and 30 opened and returned it.
216 workers completed the coreference task and 657 opened and returned it.
All but two conditions had 14 or 15 workers (the 97-98\%, 5,000+ case for coreference had 13 and the 97-98\%, 2,500-5,000 case for coreference had 12).

\subsection{Results}

\begin{figure*}
    \centering
    \includegraphics[width=0.95\linewidth]{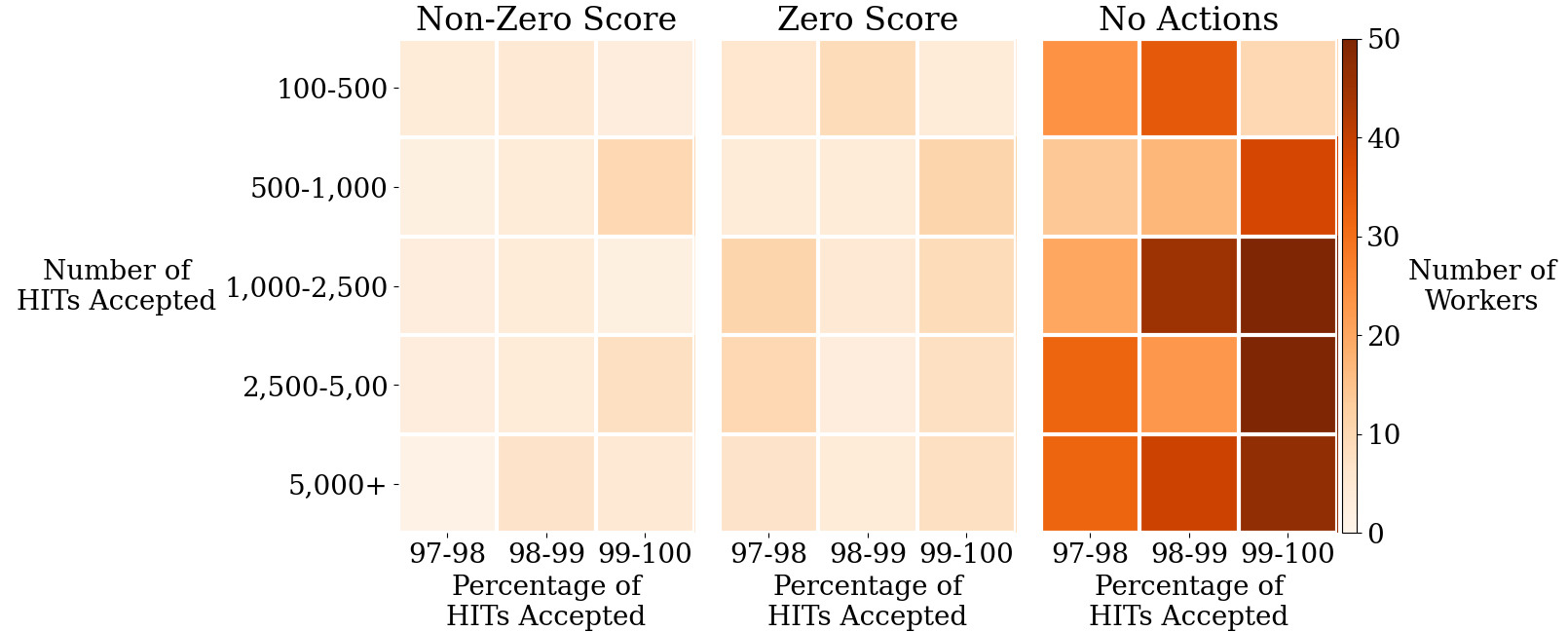}
    \caption{\label{fig:return}
    For coreference resolution, 657 workers opened and returned the HIT without completing it.
    These three heatmaps show the number of workers who:
    left partially correct annotations (Non-Zero Score),
    left entirely incorrect annotations (Zero Score),
    did not interact with the page (No Action).
    We do not include plots for sentiment analysis because only 30 workers opened and returned the HIT.
    }
\end{figure*}

Figure~\ref{fig:acc} shows the percentage of workers scoring 80 or higher on the coreference resolution task.
When the acceptance percentage is below 99, results are consistently poor, with fewer than 25\% of workers scoring above 80.
When the acceptance percentage is 99-100, groups with higher approved HITs have better scores.
Figure~\ref{fig:return} shows the number of workers who returned the HIT.\footnote{'Returning' a task means a worker choose to stop working on it, receives no pay, but also receives no penalty in their profile for failing to complete the task.}
The number of workers returning the HIT is higher in the groups with higher performance (see the last column of the rightmost plot), indicating that workers are self-selecting out.

In a follow up experiment with constraints of 99-100\% and 1,000+ using a relatively new requester account, 60 out of 92 workers scored 80 or above (65\%), indicating that there are more workers in the higher approved HITs groups.

Figure~\ref{fig:acc} also shows results for the sentiment task.
First, note that many more workers did well on the task.
Comparing the left and right, the trend for percentage of HITs accepted is repeated, with consistently poor performance from workers with values below 99\% (the left two columns).
While the best result is the same in both cases (the bottom-right), the trend in the third column is somewhat different.
Rather than a steady increase in performance as the approved HITs threshold increases, there is a U-shaped pattern.
This shows that the pattern is somewhat task dependent.

Overall, these results suggest that if the minimum approved HITs qualification is used it can be set to 2,500 without a substantial impact on work quality.
The impact of even lower values depends on the task.
In both tasks, the percentage HITs accepted qualification had a clear impact, with substantial decreases in quality from workers with a value below 99\%.
While that qualification does not directly force workers to do a substantial amount of work, it can be impacted by requesters who unfairly reject work.
Our results also suggest that simply paying workers more will not lead to better work, as the sentiment analysis task paid considerably better and did not solve the issue.

\section{Ethics and Impact Statement}

This work involved consideration of several potential impacts.
In terms of privacy, all data from workers is aggregated for the purpose of presenting results, and information from worker discussions were only sourced from publicly shared content.
In terms of payment, we estimated the effort involved and aimed to pay workers \$12 USD an hour.
See the main text for worker reported values of hourly earnings on the two tasks.
This was approved by the Michigan IRB under study ID HUM00155689.
One potential harm of this work is that it may encourage higher values of the Percentage of HITs Accepted qualification, making workers more vulnerable to requesters who unfairly reject work.

\section{Conclusion and Recommendations}

This paper identifies the issue of Qualification Labour: the implied labour created by the qualifications we define.
Based on a range of sources, we found that 5,000 approved tasks is one common threshold.
That takes approximately two months to achieve and the tasks are poorly paid.
We conducted a study of two tasks to understand how work quality correlates with these qualifications.
We found that trends are task dependent, but lower thresholds can often be used.

We recommend either not using the "HITs accepted" qualification, or running preliminary tests to identify the lowest suitable threshold for your task.
This calibration is necessary because worker performance depends on many factors, including the task type, data (including which language), user interface, and instructions.
One particularly promising method is to use controlled crowdsourcing \cite{roit-etal-2020-controlled} with a low threshold: run a short task with low or no qualifications to identify workers, then for the full task only allow those workers to participate.
This reduces the burden on workers while maintaining high quality work.

\section*{Acknowledgements}

We would like to thank Judy Kay, Ellen Stuart, Greg Durrett, and attendees at the Conference on Human Computation and Crowdsourcing for helpful feedback on earlier drafts of this paper, and the ACL reviewers for their helpful suggestions. 
This material is based in part on work supported by DARPA (grant \#D19AP00079), Bloomberg (Data Science Research Grant), and the Allen Institute for AI (Key Scientific Challenges Program).

\bibliographystyle{acl_natbib}
\bibliography{acl21fair-work}

\end{document}